\PassOptionsToPackage{dvipsnames}{xcolor}
\documentclass[letterpaper, 10 pt, conference]{ieeeconf}  % Comment this line out if you need a4paper

\IEEEoverridecommandlockouts                              % This command is only needed if 
\usepackage{cite}
\usepackage{amsmath,amssymb,amsfonts}
\usepackage{algorithmic}
\usepackage{textcomp}
\usepackage{graphicx}
\usepackage{xcolor}
\usepackage{makecell}
\usepackage{multirow}
\usepackage{booktabs}
\usepackage{float}
\usepackage{svg}
\usepackage{stfloats}
\usepackage{hyperref}
\definecolor{cvprblue}{rgb}{0.21,0.49,0.74}
\renewcommand{\baselinestretch}{0.98}
\definecolor{darkpastelgreen}{rgb}{0.01, 0.75, 0.24}
\definecolor{darkgreen}{rgb}{0.00, 0.8, 0.2}
\definecolor{darkyellow}{rgb}{0.96, 0.75, 0.00}

\let\labelindent\relax
\usepackage{enumitem}

\definecolor{mycolor}{RGB}{100, 150, 200} % 自定义颜色
\title{\LARGE \bf DiffSemanticFusion: Semantic Raster BEV Fusion for Autonomous Driving via Online HD Map Diffusion
}

\begin{document}

\author{
    Zhigang Sun$^{1*}$,
    Yiru Wang$^{1,6*\dagger}$,
    Anqing Jiang$^{1*}$,
    Shuo Wang$^{1}$, 
    Yu Gao$^{1}$, 
    Yuwen Heng$^{1}$, \\ 
    Shouyi Zhang$^{1}$,
    An He$^{1}$, 
    Hao Jiang$^{3}$, 
    Jinhao Chai$^{2}$,
    Zichong Gu$^{2}$,
    Wang Jijun$^{4}$,\\
    Shichen Tang$^{1}$, 
    Lavdim Halilaj$^{5}$,
    Juergen Luettin$^{5}$,
    Hao Sun$^{1}$
% <-this % stops a space
% \thanks{*This work was not supported by any organization}% <-this % stops a space
    \thanks{$^{1}$
            Zhigang Sun,
            Yiru Wang,
            Anqing Jiang,
            Shuo Wang,
            Yu Gao,
            Yuwen Heng,
            Shouyi Zhang,
            An He,
            Shichen Tang,
            Hao Sun are with Bosch Corporate Research, Bosch (China) Investment Ltd., Shanghai, China. * means equal contribution }
            % {\tt\small anqing.jiang\makeatletter cn.bosch.com}}
    % \thanks{$^{2}$Bernard D. Researcheris with the Department of Electrical Engineering, Wright State University,
    %         Dayton, OH 45435, USA
    %         {\tt\small b.d.researcher@ieee.org}}%
    % }
    \thanks{$^{2}$
        Jinhao Chai, Zichong Gu
        are with School of Communication and Information Engineering, Shanghai University, Shanghai, China
    }
    \thanks{$^{3}$
        Hao Jiang 
        is with Shanghai Jiaotong University, Shanghai, China
    }
    \thanks{$^{4}$ 
        Wang Jijun,
        is with AIR, Tsinghua University, Beijing, China
    }
    \thanks{$^{5}$
        Lavdim Halilaj, Juergen Luettin 
        are with Robert Bosch GmbH
    }
    \thanks{$^{6}$
        $\dagger$ represents corresponding author, {\tt\small yiru.wang@cn.bosch.com}
    }
    % \thanks{$^{6}$ 
    %     Osamu Yoshie,
    %     is with IPS, Waseda University, Tokyo, Japan
    % }
    % \thanks{$^{7}$ 
    %     Pu Jian,
    %     is with , Fudan University, Shanghai, China
    % }
}

\maketitle
\pagestyle{empty}

%%%%%%%%%%%%%%%%%%%%%%%%%%%%%%%%%%%%%%%%%%%%%%%%%%%%%%%%%%%%%%%%%%%%%%%%%%%%%%%%
\begin{abstract}
Autonomous driving requires accurate scene understanding, including road geometry, traffic agents, and their semantic relationships. In online HD map generation scenarios, raster-based representations are well-suited to vision models but lack geometric precision, while graph-based representations retain structural detail but become unstable without precise maps. To harness the complementary strengths of both, we propose DiffSemanticFusion—a fusion framework for multimodal trajectory prediction and planning. Our approach reasons over a semantic raster–fused BEV space, enhanced by a map diffusion module that improves both the stability and expressiveness of online HD map representations. We validate our framework on two downstream tasks: trajectory prediction and planning-oriented end-to-end autonomous driving. Experiments on real-world autonomous driving benchmarks, nuScenes and NAVSIM, demonstrate improved performance over several state-of-the-art (SOTA) methods. For the prediction task on nuScenes, we integrate DiffSemanticFusion with the online HD map informed QCNet, achieving a 5.1\% performance improvement. For end-to-end autonomous driving in NAVSIM, DiffSemanticFusion achieves SOTA results, with a 15\% performance gain in NavHard scenarios. In addition, extensive ablation and sensitivity studies show that our map diffusion module can be seamlessly integrated into other vector-based approaches to enhance performance. All artifacts are available at: \url{https://github.com/SunZhigang7/DiffSemanticFusion}

% 逻辑： 自动驾驶需要场景理解 -> 场景理解raster/graph -> 在在线生成地图的场景下各自优缺点 -> fusion -> map diffusion 增加稳定性 -> 下游任务是prediction以及端到端自动驾驶的planning

\end{abstract}

%%%%%%%%%%%%%%%%%%%%%%%%%%%%%%%%%%%%%%%%%%%%%%%%%%%%%%%%%%%%%%%%%%%%%%%%%%%%%%%%
\section{INTRODUCTION}
\vspace{-1pt}
End-to-end autonomous driving has attracted increasing attention due to recent progress in  learning based perception and planning systems, including object detection~\cite{huang2021bevdet,li2024bevformer,wang2022detr3d}, multi-object tracking~\cite{zeng2022motr,zhang2022bytetrack}, online mapping~\cite{liao2022maptr,liao2025maptrv2,yuan2024streammapnet,jiang2025sparsemext}, prediction~\cite{Cui2018MultimodalTP,PhanMinh2019CoverNetMB,Li2020EvolveGraphMT,Gao2020VectorNetEH} and planning~\cite{huang2023gameformer,cheng2024pluto,zheng2025diffusion}. These advances have enabled the learning of driving policies directly from raw sensor inputs, bypassing the need for hand-crafted rules. 

Two-stage end-to-end frameworks decompose the driving task into intermediate perception and planning modules, typically leveraging structured representations such as bird’s-eye view (BEV) maps or object lists. This design enhances interpretability and training stability, but may suffer from sub-optimal performance due to error propagation between stages and limited capacity for joint optimization. While map-informed prediction and planning have shown notable benefits by using high-definition (HD) maps to provide structural context, their effectiveness often diminishes in scenarios where online maps are incomplete, noisy, or misaligned. Such limitations can significantly affect downstream tasks like motion prediction and planning, particularly in dynamic or previously unseen environments. Moreover, strict reliance on pre-defined map structures can hinder generalization and robustness in real-world where map inaccuracies are inevitable.
\begin{figure}[t]
    \centering
    \includegraphics[width=0.48\textwidth]
    {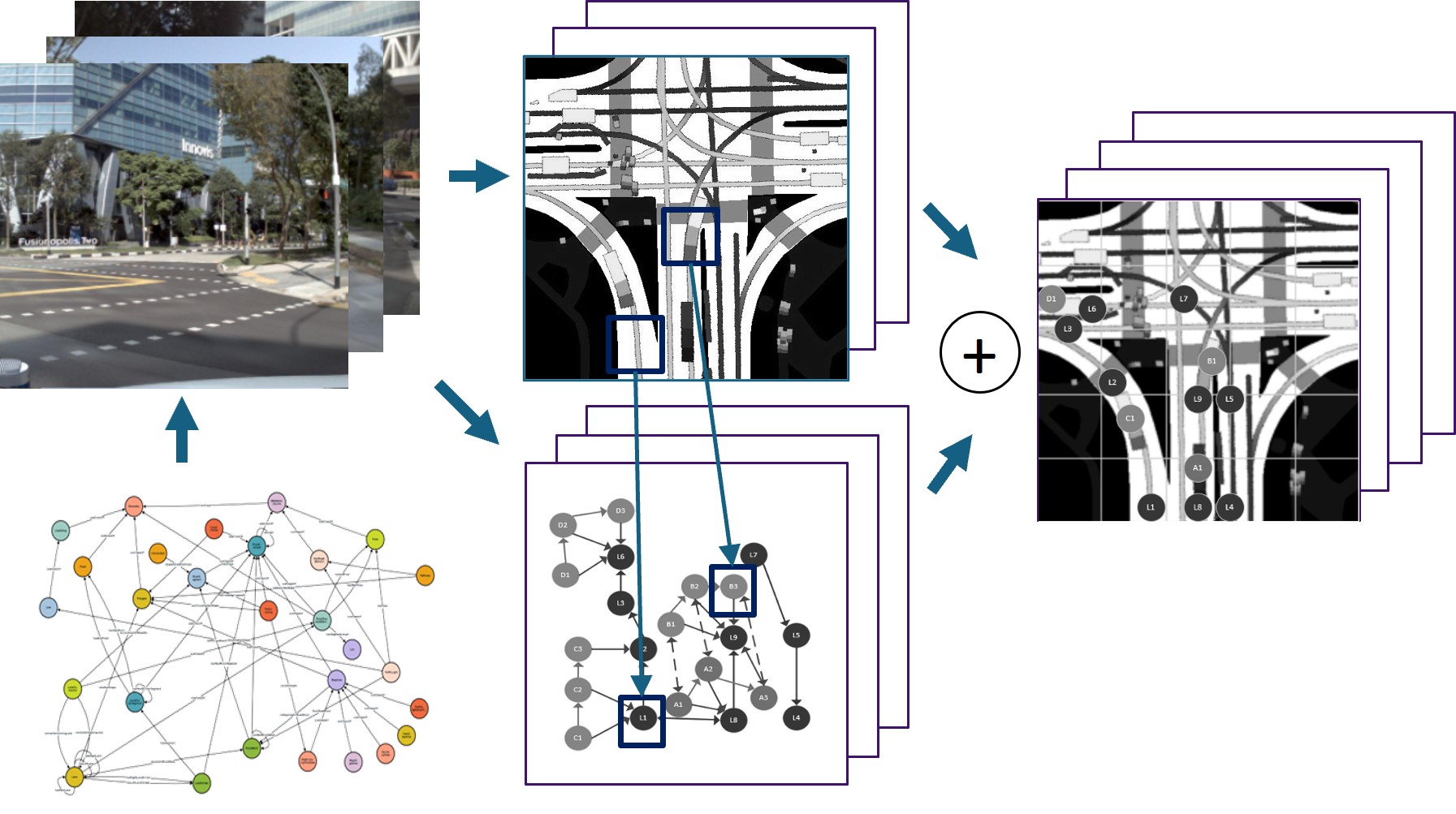} 
    \vspace{-18pt}
    \caption{Semantic Raster Image Fusion}
    \label{Overview}
    \vspace{-18pt}
\end{figure}
To address these challenges, we propose online HD map diffusion, a novel framework that enhances the robustness of map-informed prediction under online mapping conditions. Rather than treating the map as a fixed input, map diffusion introduces a learnable mechanism that iteratively refines and denoises online map representations via a diffusion-based generative process. This approach allows the model to adaptively recover reliable map features, even in the presence of uncertainty or noise, thereby improving the quality of trajectory prediction and downstream planning. By bridging the gap between static HD maps and dynamic online mapping, online HD map diffusion offers a resilient and scalable solution for safe and efficient autonomous driving.

In contrast, one-stage end-to-end approaches~\cite{hu2023planning,jiang2023vad,chen2024vadv2,sun2024sparsedrive,jiang2025diffvla} directly regress control commands or trajectories from raw sensor inputs, enabling full-system differentiability and potentially higher performance. However, they often face challenges related to interpretability, data efficiency, and deployment in safety-critical settings. To combine the advantages of both paradigms, we propose a semantic raster BEV fusion framework that incorporates intermediate representations within an end-to-end trainable architecture. This design retains the interpretability and structure of two-stage pipelines while enabling the global optimization benefits of one-stage systems. Such integration has shown promise in achieving both high performance and reliable decision-making in complex driving scenarios.

% \vspace{-5pt}
\textbf{The main contributions are:}
\begin{itemize}
    \item We propose the online HD map diffusion module, which improves the robustness of online HD maps under noisy or incomplete conditions, enhancing downstream prediction and planning performance.
    \item We design a semantic raster BEV fusion architecture that combines raster-based, graph-based map and BEV-feature representations in BEV space, leveraging their complementary strengths for more stable and informative scene understanding.
    \item We validate our method on two tasks: prediction on nuScenes~\cite{caesar2020nuscenes} and planning oriented end-to-end autonomous driving in NAVSIM~\cite{dauner2024navsim}. Both achieve SOTA performance. DiffSemanticFusion improves online HD map informed QCNet~\cite{zhou2023query} prediction by 5.1\% and achieves a 15\% gain in NAVSIM Navhard planning scenarios. Extensive studies also show strong compatibility with other vector-based methods.
\end{itemize}

\section{Related Work}
\begin{figure*}[t]
    \centering
\includegraphics[width=\textwidth]{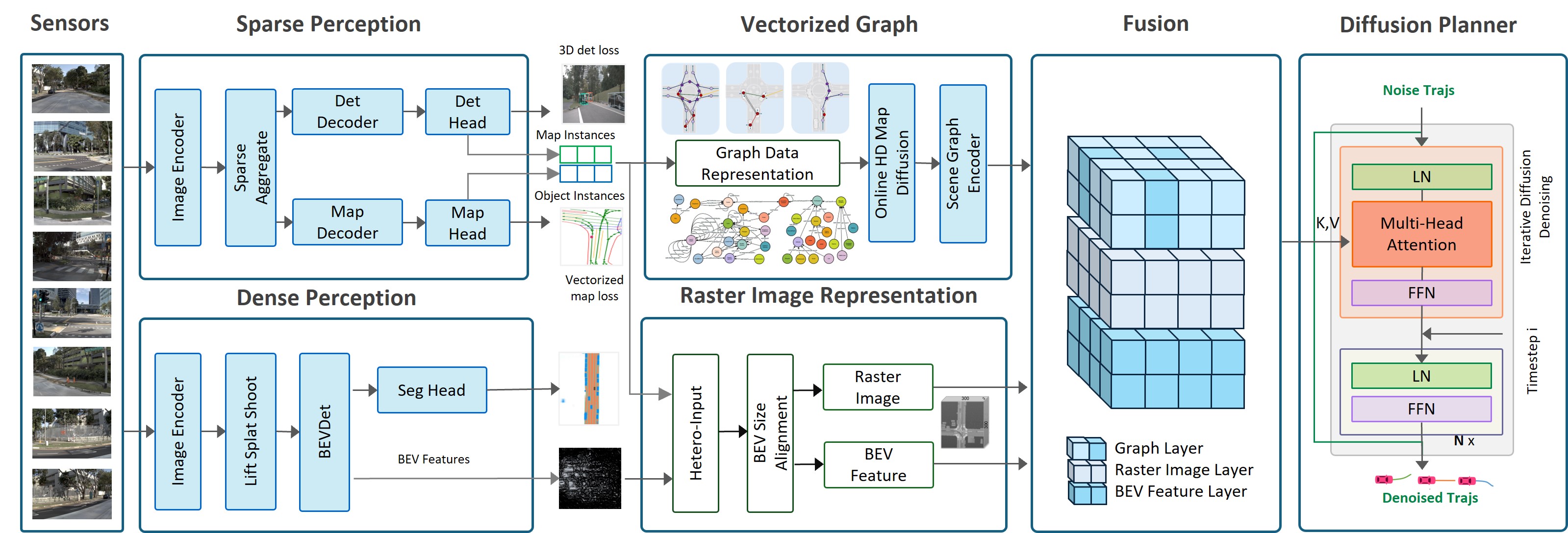} % Replace 'image_file_name' with the actual file name of your image
    \vspace{-15pt}
    \caption{DiffSemanticFusion Overview: \textit{Sparse Perception} utilizes \textit{Sparse4D}~\cite{lin2022sparse4d} to extract bounding boxes of dynamic objects and map elements. \textit{Dense Perception} utilizes \textit{LSS}~\cite{philion2020lift} and \textit{BEVDet}~\cite{huang2021bevdet} to extract BEV features. \textit{Vectorized Graph} utilizes \textit{SemanticFormer}~\cite{sun2024semanticformer} to extract graph information. \textit{Fusion} fuses heterogeneous representations to unified space and \textit{Diffusion Planner} serves as trajectories decoder.}
    \vspace{-15pt}
    \label{fig:diffusionfusion}
\end{figure*}
\subsection{Traffic Scene Representation}
Raster-based representation approaches encode the entire traffic scene into bird's-eye-view (BEV) images with multiple channels, where each channel represents different elements of the scenes~\cite{Cui2018MultimodalTP,PhanMinh2019CoverNetMB,Berkemeyer2021FeasibleAA}. On top of these raster representations, convolutional neural networks (CNNs) are typically employed to extract features from the scene. The learned representation is then passed through fully connected layers to generate trajectory predictions. However, a major limitation of raster-based models is their reliance on raw pixel-level input, which lacks high-level semantic understanding and requires networks to infer complex relationships directly from image data. In contrast, graph-based representation methods model traffic scenes as collections of vectors, polylines, or graphs~\cite{Gao2020VectorNetEH,Liang2020Learning,Casas2019SpAGNNSG,Liu2023LAformerTP}. These approaches operate at a higher level of abstraction, removing the need to learn from low-level pixels and thereby enhancing robustness to scene variations. For example, VectorNet~\cite{Gao2020VectorNetEH} encodes map features and agent trajectories as polylines and models their interactions using a global graph. However, many graph-based models are limited to homogeneous graphs, representing only a single type of entity and relation. To overcome this, recent works have introduced heterogeneous graph representations, which incorporate diverse entity types (e.g., vehicles, bicycles, pedestrians) and relation types (e.g., agent-to-lane, agent-to-agent)~\cite{Mo2022MultiAgentTP,Jia2022HDGTHD,Monninger2023SCENE,mlodzian2023nuscenes,wang2024socialformer,sun2024semanticformer}. For instance, SemanticFormer~\cite{sun2024semanticformer} generates multimodal trajectories by reasoning over a semantic traffic scene graph using attention mechanisms and graph neural networks. While powerful, such approaches often involve complex graph construction and incur significant computational overhead. 

To address these challenges, we propose DiffSemanticFusion, which combines the strengths of both raster- and graph-based representations by fusing features in BEV space. This hybrid approach retains the advantages of explicit geometric structure and global semantic context, offering a more expressive and efficient solution for trajectory prediction.
\vspace{-2pt}
\subsection{Diffusion Mechanism in Autonomous Driving}
\vspace{-2pt}
Diffusion models, initially proposed for image generation tasks~\cite{ho2020denoising}, have recently been extended to a broad spectrum of domains beyond computer vision. In the field of autonomous driving, they have demonstrated notable effectiveness in trajectory prediction~\cite{jiang2023motiondiffuser, zhong2022guided}, motion planning~\cite{zheng2025diffusion}, and end-to-end autonomous driving~\cite{liao2025diffusiondrive}. MotionDiffuser\cite{jiang2023motiondiffuser} adopts a conditional diffusion model to generate target trajectories from Gaussian noise. Diffusion Planner\cite{zheng2025diffusion} employs a customized diffusion-based architecture to enable high-performance motion planning, while mitigating the dependence on hand-crafted rules. DiffusionDrive~\cite{liao2025diffusiondrive} further advances this line of work by introducing a truncated diffusion policy and an efficient decoding mechanism for real-time, end-to-end autonomous driving.

While these approaches leverage diffusion models primarily for direct trajectory generation, our work explores a different dimension. To the best of our knowledge, we are the first to propose an online HD map diffusion module from the perspective of prediction and planning. This module is designed to enhance the stability and consistency of online HD maps, which are critical for robust long-term autonomy.

\subsection{End-to-End Autonomous Driving}
Previous end-to-end autonomous driving approaches typically rely on a single scene representation. UniAD~\cite{hu2023planning} pioneers the integration of multiple perception tasks to enhance planning performance within a unified framework. VAD~\cite{jiang2023vad} introduces compact vectorized scene representations to improve computational efficiency. Building upon this, a line of work~\cite{chitta2022transfuser,li2025end,liao2025diffusiondrive,li2025hydra,yao2025drivesuprim} adopts the single-trajectory prediction paradigm for improved planning accuracy. More recently, VADv2~\cite{chen2024vadv2} explores multi-mode planning by scoring and selecting from a fixed trajectory set, while Hydra-MDP~\cite{li2024hydra} enhances this approach by incorporating rule-based supervision. SparseDrive~\cite{sun2024sparsedrive} investigates an alternative, BEV-free pipeline. DiffusionDrive~\cite{liao2025diffusiondrive} proposes a generative framework leveraging diffusion models for diverse and high-quality trajectory generation. Despite these advances, existing methods predominantly rely on either dense BEV or sparse representations. In contrast, we present the first end-to-end autonomous driving framework that unifies sparse scene representations with dense BEV features, enabling complementary strengths of both modalities. Our approach demonstrates superior planning performance by leveraging the structured abstraction of sparse and dense BEV representations in a unified architecture.

\section{Method}
\subsection{Preliminary}
\subsubsection{Problem Formulation}
Trajectory prediction in modular autonomous driving focuses on forecasting the future motion of surrounding agents (e.g., vehicles, pedestrians, cyclists) is predicted based on their past observed trajectories. At the current time step $t=0$, the observed trajectory of an agent $i$ over the past $T_h$ steps is represented as $\tau_{\text{obs}}^i = \left\{(x_t^i, y_t^i)\right\}_{t = -T_h + 1}^{0}$. The goal is to predict the future trajectory over the next $T_f$ steps, $\tau_{\text{pred}}^i = \left\{(x_t^i, y_t^i)\right\}_{t = 1}^{T_f}$. All trajectories are represented in a consistent coordinate system, such as the map coordinate system or an agent-centric coordinate system. The prediction model may output one or more trajectory hypotheses per agent to account for the uncertainty and multimodal nature of real-world behaviors. Planning-oriented end-to-end autonomous driving takes raw sensor data as input and predicts the future trajectory of the ego-vehicle. The predicted trajectory is represented as a sequence of waypoints, $\tau_{\text{ego}} = \left\{(x_t, y_t)\right\}_{t = 1}^{T_f}$, where $T_f$ denotes the planning horizon, and $(x_t, y_t)$ is the location of each waypoint at time $t$, represented in the current ego-vehicle coordinate system.
\subsubsection{Conditional Diffusion Model}
The conditional diffusion framework models the generation process by introducing noise to a clean data sample through a forward stochastic process. At each diffusion step $i$, the perturbed sample $\tau^i$ is generated from the original sample $\tau^0$ by equation~\ref{eq_diffusion_forward}.
\begin{equation}
    q\left(\tau^i \mid \tau^0\right) = \mathcal{N}\left(\tau^i ; \sqrt{\bar{\alpha}_i} \, \tau^0, \left(1 - \bar{\alpha}_i\right)\mathbf{I}\right)
\label{eq_diffusion_forward}
\end{equation}
where the superscript $i$ denotes the diffusion timestep, $\tau^0$ is the clean (ground-truth) data, and $\bar{\alpha}_i = \prod_{s=1}^i \alpha_s = \prod_{s=1}^i (1 - \beta_s)$ is the accumulated product of noise coefficients. Here, $\beta_s$ represents the predefined noise schedule. To recover the clean sample from the noisy observation, we train a reverse process model $f_\theta(\tau^i, z, i)$, where $\theta$ are the learnable parameters and $z$ denotes the conditioning input. The model aims to estimate the original sample $\tau^0$ conditioned on both $\tau^i$ and $z$. During inference, we begin with an initial sample $\tau^T$ drawn from a Gaussian prior and iteratively denoise it using the learned reverse transitions. The overall sampling objective conditioned on $z$ is expressed as equation~\ref{eq_diffusion_sampling}.
\begin{equation}
    p_\theta(\tau^0 \mid z) = \int p(\tau^T) \prod_{i=1}^{T} p_\theta(\tau^{i-1} \mid \tau^i, z) \, \mathrm{d}\tau^{1:T}
\label{eq_diffusion_sampling}
\end{equation}
\vspace{-10pt}
\begin{figure}[t]
  \centering
  \includegraphics[width=0.45\textwidth]{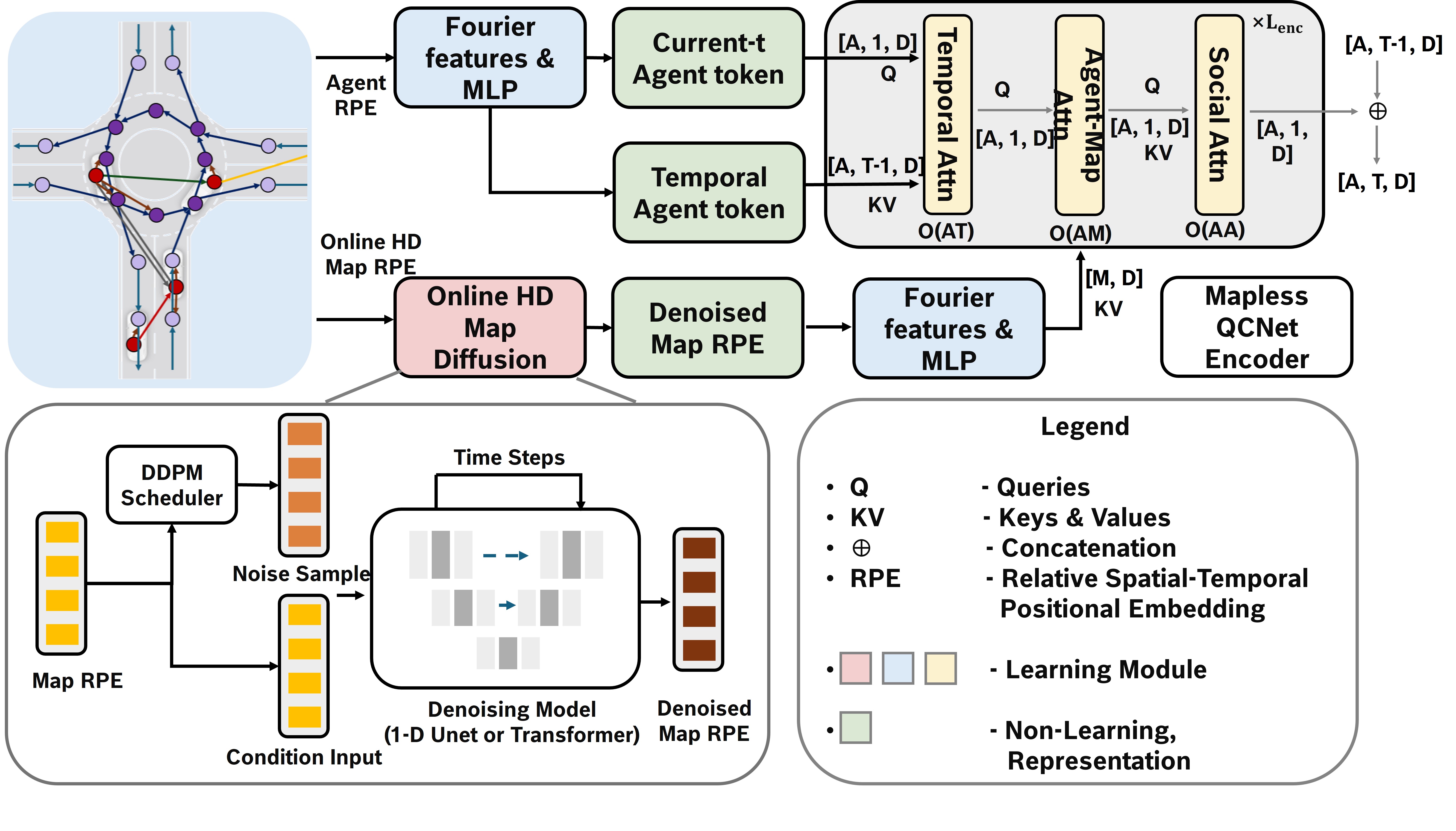} 
  \vspace{-10pt}
  \caption{Mapless QCNet Encoder with Online HD Map Diffusion}
    \vspace{-15pt}
  \label{fig:map_diffusion}
\end{figure}
\subsection{Online HD Map Diffusion Module}
After online map perception, we can get the map information like divider, boundary, pedestrian crossing represented by an ordered sequence of sparse lane vectors as $\mathbf{L}_{1: N}=\left\{v_1, v_2, \ldots, v_N\right\}$, where $N$ denotes the total vector length. Each lane vector $v_n=\left[d_{n, s}, d_{n, e}, a_n\right]$  where $d_{n, s}$, $d_{n, e}$ denote the start and end points, respectively, and $a_n$ corresponds to lane point $n$’s attribute features, such as orientation, polyline type, etc. We first construct the diffusion process by adding Gaussian noise to sparse lane vectors $\{v_k\}_{k=1}^{N}$ on the training set.
We truncate the diffusion noise schedule to diffuse the lane vectors to the lane vector based Gaussian distribution as equation~\ref{eq_diffusion_map}.
\begin{equation}
    \tau_k^i = \sqrt{\bar{\alpha}^i}v_k + \sqrt{1-\bar{\alpha}^i}\boldsymbol{\epsilon}, \quad \boldsymbol{\epsilon} \sim \mathcal{N}(0, \mathbf{I})
\label{eq_diffusion_map}
\end{equation}
where $i \in [1,T_\text{trunc}]$ and $T_\text{trunc} \ll T$ is the truncated diffusion steps. During training, the diffusion decoder $f_\theta$ takes as input all $N$ noisy sparse lane vectors $\{\tau_k^i\}_{k=1}^{N}$ and predicts denoised sparse lane vectors $\{\hat{\tau}_k\}_{k=1}^{N}$ as equation~\ref{eq_diffusion_map_1}.
\begin{equation} 
    \{\hat{\tau}_k\}_{k=1}^{N} = f_\theta(\{\tau_k^i\}_{k=1}^{N}, z)
\label{eq_diffusion_map_1}
\end{equation}
where $z$ represents the conditional information. We calculate the MSELoss between all the denoised ordered sequence of sparse lane vectors $\{\hat{\tau}_k\}_{k=1}^{N}$ and ground truth $\{\tau_{gt}\}_{k=1}^{N}$. Online map will generate $N^\text{lane}$ ordered sequence of sparse lane vectors, then online map diffusion training objective is shown in equation~\ref{eq_diffusion_map_loss}, $m$ represents vector in $m$'s lane.
\begin{equation}
    \mathcal{L} = \sum_{m=1}^{N_\text{lane}}\sum_{k=1}^{N} \mathcal{L}_{\text{MSE}}(\hat{\tau}_{km}, \tau_\text{gt})
\label{eq_diffusion_map_loss}
\end{equation}
Online map diffusion module is agnostic to the coordinate system and can be seamlessly integrated into any vector-based online HD map, whether the vectors are represented in cartesian or polar coordinates.

\begin{table*}[htbp]
\normalsize
\scriptsize
\caption{\textbf{Performance on the NAVSIM-V2 Navtest Benchmark for Planning oriented End-to-End Autonomous Driving.}}
\centering
\begin{tabular}{l|l|l| l l l l l l l l l |l}
    \toprule
    Method 
    & Backbone
    & Sensor
    & $\text{NC$\uparrow$}$
    & $\text{DAC$\uparrow$}$
    & $\text{DDC$\uparrow$}$
    & $\text{TLC$\uparrow$}$
    & $\text{EP$\uparrow$}$
    & $\text{TTC$\uparrow$}$
    & $\text{LK$\uparrow$}$ 
    & $\text{HC$\uparrow$}$ 
    & $\text{EC$\uparrow$}$ 
    & $\text{EPDMS$\uparrow$}$  \\
    \midrule
    Human Agent  & - &  - & \makecell{100} & \makecell{100} & \makecell{99.8} & \makecell{100} & \makecell{87.4} & \makecell{100} & \makecell{100} & \makecell{98.1} & \makecell{90.1} & 90.3 \\
    Ego Status MLP  & - &  camera & \makecell{93.1} & \makecell{77.9} & \makecell{92.7} & \makecell{99.6} & \makecell{86.0} & \makecell{91.5} & \makecell{89.4} & \makecell{98.3} & \makecell{85.4} & 64.0 \\
    \midrule
    Transfuser  & ResNet34 & camera & \makecell{96.9} & \makecell{89.9} & \makecell{97.8} & \makecell{99.7} & \makecell{87.1} & \makecell{95.4} & \makecell{92.7} & \makecell{98.3} & \makecell{87.2} & 76.7 \\
    Diffusiondrive~\cite{liao2025diffusiondrive}  & ResNet34 & camera & \makecell{98.0} & \makecell{96.0} & \makecell{99.5} & \makecell{99.8} & \makecell{87.7} & \makecell{97.1} & \makecell{97.2} & \makecell{98.3} & \makecell{87.6} & 84.3 \\
    HydraMDP++~\cite{li2025hydra}  & ResNet34 & camera & \makecell{97.2} & \makecell{97.5} & \makecell{99.4} & \makecell{99.6} & \makecell{83.1} & \makecell{96.5} & \makecell{94.4} & \makecell{98.2} & \makecell{70.9} & 81.4\\
    DriveSuprim~\cite{yao2025drivesuprim}  & ResNet34 & camera & \makecell{97.5} & \makecell{96.5} & \makecell{99.4} & \makecell{99.6} & \makecell{88.4} & \makecell{96.6} & \makecell{95.5} & \makecell{98.3} & \makecell{77.0} & 83.1 \\
    \textbf{DiffSemanticFusion}  & ResNet34 & camera & \makecell{98.4} & \makecell{96.3} & \makecell{99.5} & \makecell{99.8} & \makecell{88.5} & \makecell{97.6} & \makecell{97.5} & \makecell{98.4} & \makecell{87.7} & \textbf{85.1} \\
    \midrule
    WOTE~\cite{li2025end} & V2-99 & camera & \makecell{98.5} & \makecell{96.8} & \makecell{98.8} & \makecell{99.8} & \makecell{86.1} & \makecell{97.9} & \makecell{95.5} & \makecell{98.3} & \makecell{82.9} & 84.2 \\
    Diffusiondrive~\cite{liao2025diffusiondrive} & V2-99 & camera & \makecell{98.2} & \makecell{96.3} & \makecell{99.6} & \makecell{99.8} & \makecell{87.5} & \makecell{97.5} & \makecell{97.1} & \makecell{98.3} & \makecell{87.7} & 85.0 \\
    HydraMDP++~\cite{li2025hydra} & V2-99 & camera & \makecell{98.4} & \makecell{98.0} & \makecell{99.4} & \makecell{99.8} & \makecell{87.5} & \makecell{97.7} & \makecell{95.3} & \makecell{98.3} & \makecell{77.4} & 85.1 \\
    DriveSuperim~\cite{yao2025drivesuprim} & V2-99 & camera & \makecell{97.8} & \makecell{97.9} & \makecell{99.5} & \makecell{99.9} & \makecell{90.6} & \makecell{97.1} & \makecell{96.6} & \makecell{98.3} & \makecell{77.9} & 86.0 \\
    \textbf{DiffSemanticFusion}  & V2-99 & camera & \makecell{98.5} & \makecell{97.4} & \makecell{99.5} & \makecell{99.8} & \makecell{88.6} & \makecell{97.5} & \makecell{97.7} & \makecell{98.3} & \makecell{88.0} & \textbf{86.5} \\
\bottomrule
\end{tabular}
% \end{adjustbox}
\label{table:result_navtest}
\end{table*}
\vspace{-5pt}
\begin{table*}[htbp]
\normalsize
\scriptsize
\caption{\textbf{Performance on the NAVSIM-V2 Navhard Benchmark for Planning oriented End-to-End Autonomous Driving.}}
\centering
\begin{tabular}{l|l|l| l l l l l l l l l |l}
    \toprule
    Method 
    & Backbone
    & Stage
    & $\text{NC$\uparrow$}$
    & $\text{DAC$\uparrow$}$
    & $\text{DDC$\uparrow$}$
    & $\text{TLC$\uparrow$}$
    & $\text{EP$\uparrow$}$
    & $\text{TTC$\uparrow$}$
    & $\text{LK$\uparrow$}$ 
    & $\text{HC$\uparrow$}$ 
    & $\text{EC$\uparrow$}$ 
    & $\text{EPDMS$\uparrow$}$  \\
    \midrule
    
    PDM-Closed~\cite{dauner2023parting}  & - &   \makecell{Stage 1 \\ Stage 2} & \makecell{94.4 \\ 88.1} & \makecell{98.8 \\ 90.6} & \makecell{100 \\ 96.3} & \makecell{99.5 \\ 98.5} & \makecell{100 \\ 100} & \makecell{93.5 \\ 83.1 } & \makecell{99.3 \\ 73.7} & \makecell{87.7 \\ 91.5} & \makecell{36.0 \\ 25.4} & 51.3    \\
    \midrule
    LTF~\cite{chitta2022transfuser} & ResNet34 &   \makecell{Stage 1 \\ Stage 2} &
    \makecell{96.2 \\ 77.7} & \makecell{79.5 \\ 70.2} & \makecell{99.1 \\ 84.2} & \makecell{99.5 \\ 98.0} & \makecell{84.1 \\ 85.1} & \makecell{95.1 \\ 75.6} & \makecell{94.2 \\ 45.4} & \makecell{97.5 \\ 95.7} & \makecell{79.1 \\ 75.9} & 23.1    \\
    \midrule
    
    DiffusionDrive~\cite{liao2025diffusiondrive} & ResNet34 &   \makecell{Stage 1 \\ Stage 2}  &
    \makecell{95.9 \\ 79.5} & \makecell{84.0 \\ 72.8} & \makecell{98.6 \\ 84.1} & \makecell{99.8 \\ 98.4} & \makecell{84.4 \\ 87.5} & \makecell{96.0 \\ 76.2} & \makecell{95.1 \\ 46.6} & \makecell{97.6 \\ 96.1} & \makecell{71.1 \\ 62.4} &  26.1  \\
    \midrule

    WOTE~\cite{li2025end} & ResNet34 &   \makecell{Stage 1 \\ Stage 2}  &
    \makecell{97.4 \\ 81.2} & \makecell{88.2 \\ 77.7 } & \makecell{97.7 \\ 84.8 } & \makecell{99.3 \\ 98.1 } & \makecell{82.7 \\ 85.9} & \makecell{96.4 \\ 78.5 } & \makecell{90.8 \\ 46.2 } & \makecell{97.3 \\ 96.6} & \makecell{68.0 \\ 63.3 } &  27.9  \\
    \midrule
    
    % \midrule
    \makecell[l]{\textbf{DiffSemanticFusion}}  & ResNet34 &   \makecell{Stage 1 \\ Stage 2}  &
    \makecell{99.1 \\ 80.5} & \makecell{88.2 \\ 73.8 } & \makecell{99.7 \\ 86.7  } & \makecell{99.5 \\ 98.2 } & \makecell{83.6 \\ 85.9} & \makecell{97.5 \\ 76.3  } & \makecell{97.1 \\ 49.0  } & \makecell{97.5 \\ 95.7 } & \makecell{76.8 \\ 62.9 } & \textbf{32.2 {\color{orange} ($+15\%$)}}     \\
    \midrule

    \makecell[l]{\textbf{DiffSemanticFusion}}  & V2-99 &   \makecell{Stage 1 \\ Stage 2}  &
    \makecell{98.6 \\ 81.8} & \makecell{90.6 \\ 75.3 } & \makecell{99.5 \\ 87.7 } & \makecell{99.5 \\ 98.0 } & \makecell{83.4 \\ 85.0 } & \makecell{97.3 \\ 78.8 } & \makecell{96.8 \\ 48.6 } & \makecell{97.5 \\ 96.1 } & \makecell{70.6 \\ 61.5} & \textbf{33.4 {\color{orange} ($+19\%$)}}    \\
\bottomrule

\end{tabular}
% \end{adjustbox}
\label{table:result_navhard}
\vspace{-14pt}
\end{table*}

\subsection{Semantic Raster Image Fusion Module}

After the upstream hybrid perception module, three distinct representations of the traffic scene are obtained: a BEV feature map $B \in \mathbb{R}^{C_1 \times H_1 \times W_1}$, an actor-specific BEV raster image $R \in \mathbb{R}^{C_2 \times H_2 \times W_2}$ and a heterogeneous traffic scene graph $G=(V, E, \tau, \phi)$, which has nodes $v \in V$, their types $\tau(v)$, and edges $(u, v) \in E$, with edge types $\phi(u, v)$. To effectively exploit the complementary characteristics of these heterogeneous representations, different learning strategies are proposed to map them into a unified space $U\in \mathbb{R}^{C \times H \times W}$. For the BEV feature, we adopt the BEV feature projection method from BEVDet~\cite{huang2021bevdet} to implicitly map raw sensor data into unified space $U$. For BEV raster image $R$, any CNN architecture can be used as the base network to map the original image to unified space $U$. Following the~\cite{Cui2018MultimodalTP}, we use MobileNet-v2~\cite{sandler2018mobilenetv2}. For heterogeneous traffic scene graph $G$, we utilize the Semanticformer~\cite{sun2024semanticformer} to map the complex graph to hidden space $Z \in \mathbb{R}^{C}$. Then the graph nodes are projected onto the unified space $U$ based on their geometric coordinates $(x, y, z)$ by applying a perspective-to-top-down transformation, enabling spatially consistent reasoning in the planar domain. Simply concatenating the graph nodes with the unified spatial representation $U$ often leads to highly sparse feature maps, where most spatial locations contain little to no meaningful information. To mitigate this, we aggregate the graph node features into $U$ through channel-wise addition, promoting denser feature propagation while preserving the underlying geometric structure.

The Semantic raster image fusion module projects heterogeneous representations into a unified spatial space $U \in \mathbb{R}^{C \times H \times W}$, facilitating the construction of a coherent BEV feature map that effectively supports various downstream tasks such as motion prediction and planning.

\subsection{DiffSemanticFusion Architecture}
We employ hybrid perception module to extract sparse and dense representations of traffic scene as shown in figure~\ref{fig:diffusionfusion}. Dense perception module utilizes LSS~\cite{philion2020lift} and BEVDet~\cite{huang2021bevdet} to extract BEV features $B \in \mathbb{R}^{C_1 \times H_1 \times W_1}$. And sparse perception module utilizes Sparse4D~\cite{lin2022sparse4d} to extract bounding boxes of dynamic objects and map elements as shown in equation~\ref{eq_od} and ~\ref{eq_map}. 
\begin{equation} 
    \left\{ x, y, z, \ln w, \ln h, \ln l, \sin{yaw}, \cos{yaw}, vx, vy, vz\right\}
\label{eq_od}
\end{equation}
\vspace{-14pt}
\begin{equation} 
    \left\{x_{0},y_{0},x_{1},y_{1},...,x_{N_p-1},y_{N_p-1} \right\}
\label{eq_map}
\end{equation}
The vectorized graph module encodes structured semantic information, which is first refined and completed using an online HD map diffusion module. Then, following the VectorNet~\cite{Gao2020VectorNetEH} approach, the elements are represented as a fully connected graph where information is propagated according to the geometric relationships of traffic elements in the scene. To further enhance this representation, we adopt the SemanticFormer~\cite{sun2024semanticformer} framework to perform heterogeneous graph learning, projecting the structured traffic scene elements into a graph hidden space. The Raster Image module rasterizes the structured information into an image, which is then processed by a MobileNet-v2~\cite{sandler2018mobilenetv2} network to extract features projected into the raster image hidden space. The Fusion module combines features from the BEV representation (dense perception), graph hidden space (structured vector graph), and image hidden space (rasterized semantic map). These features are either concatenated or added before being passed to the diffusion-based planner module. The planner then decodes the final trajectory output through a denoising diffusion process.
\subsection{Motion Prediction with Online HD Map Diffusion}
To evaluate the effectiveness of our proposed online HD map diffusion module, we integrate it into the prediction backbone of QCNet~\cite{zhou2023query}, as illustrated in Figure~\ref{fig:map_diffusion}. Specifically, we replace the original HD map used in QCNet with the online HD map generated by StreamMapNet~\cite{yuan2024streammapnet}. Since the original map module in QCNet operates in polar coordinates, we incorporate the online HD map diffusion module at both the \textit{point-to-polygon} and \textit{polygon-to-polygon} stages under the polar coordinate representation. To further validate the generality of our approach, we also convert the polar map representation into Cartesian coordinates and apply the online HD map diffusion module in this domain as well. The entire model is trained in an end-to-end manner, where the diffusion loss is jointly optimized with the original QCNet loss, enabling unified training and seamless integration of the proposed module.

\begin{table*}[ht]
   \scriptsize
    \centering
    \caption{\textbf{Performance on the nuScenes Benchmark for Mapless Prediction}} 
    \begin{tabular}{l|l|l|ll|lllll}
        \toprule
        \textbf{Method Source} & \textbf{Prediction} &\textbf{Online Map} & \textbf{Diffusion}  &  \textbf{Options}  & \textbf{ADE} $\downarrow$ & \textbf{AHE} $\downarrow$ & \textbf{FDE} $\downarrow$ & \textbf{FHE} $\downarrow$ & \textbf{MR} $\downarrow$ \\
        \midrule
        SATP~\cite{dong2025leveraging} & DenseTNT~\cite{gu2021densetnt} & SD                  & -  & -  & 1.379 &   -   & 2.255 &   -  & 0.444 \\
        SATP~\cite{dong2025leveraging} & DenseTNT~\cite{gu2021densetnt} & MapTRv2~\cite{liao2025maptrv2}             & -  & -  & 1.174 &   -   & 2.191 &  -   & 0.403 \\
        SATP~\cite{dong2025leveraging} & DenseTNT~\cite{gu2021densetnt} & MapTRv2~\cite{liao2025maptrv2}  + SATP      & -  & -  & 1.017 &   -   & 1.920 &  -  &  0.379 \\
        Unc.~\cite{gu2024producing} & DenseTNT~\cite{gu2021densetnt} & StreamMapNet~\cite{yuan2024streammapnet}        & -  & -  & 0.949 &   -   & 1.740 &  -  &  0.256 \\
        Unc.~\cite{gu2024producing} & DenseTNT~\cite{gu2021densetnt} & StreamMapNet~\cite{yuan2024streammapnet} + Unc. & -  & -  & 0.903 &   -   & 1.645 &  -  &  0.235 \\
        
        \midrule
        SATP~\cite{dong2025leveraging} & HiVT~\cite{zhou2022hivt} & SD                  & - & -  & 0.454  &   -   & 0.890 &  -  & 0.107 \\
        SATP~\cite{dong2025leveraging} & HiVT~\cite{zhou2022hivt} & MapTRv2~\cite{liao2025maptrv2}              & - & -  & 0.399  &   -   & 0.832 &  -  & 0.095 \\
        SATP~\cite{dong2025leveraging} & HiVT~\cite{zhou2022hivt} & MapTRv2~\cite{liao2025maptrv2}  + SATP      & - & -  & 0.366  &   -   & 0.715 &  -  & 0.071 \\
        Unc.~\cite{gu2024producing} & HiVT~\cite{zhou2022hivt} & StreamMapNet~\cite{yuan2024streammapnet}        & - & -  & 0.397  &   -   & 0.818 &  -  & 0.092 \\
        Unc.~\cite{gu2024producing} & HiVT~\cite{zhou2022hivt} & StreamMapNet~\cite{yuan2024streammapnet} + Unc. & - & -  & 0.384  &   -   & 0.795 &  -  & 0.086 \\

        \midrule
        SATP~\cite{dong2025leveraging} & QCNet~\cite{zhou2023query} & SD                  & - & -  & 0.401  &   -   & 0.871 &  -  & 0.091 \\
        SATP~\cite{dong2025leveraging} & QCNet~\cite{zhou2023query} & MapTRv2~\cite{liao2025maptrv2}             & - & -  & 0.385  &   -   & 0.801 &  -  & 0.087 \\
        SATP~\cite{dong2025leveraging} & QCNet~\cite{zhou2023query} & MapTRv2~\cite{liao2025maptrv2}  + SATP      & - & -  & 0.362  &   -   & 0.673 &  -  & 0.069 \\ 
        \midrule
        \textbf{Ours} & QCNet~\cite{zhou2023query} & StreamMapNet~\cite{yuan2024streammapnet}   & - & -  & 0.354  & 0.324 & 0.717 & 0.407 & 0.068 \\

        \midrule
        \textbf{Ours} & QCNet~\cite{zhou2023query} & StreamMapNet~\cite{yuan2024streammapnet} & Transformer & pt2pl  & 0.340 & 0.320 & 0.699 & 0.361 & 0.0633 \\
        \textbf{Ours} & QCNet~\cite{zhou2023query} & StreamMapNet~\cite{yuan2024streammapnet} & Transformer & pl2pl  & 0.346 & 0.308 & 0.699 & 0.328 & 0.0626 \\
        \textbf{Ours} & QCNet~\cite{zhou2023query} & StreamMapNet~\cite{yuan2024streammapnet} & Transformer & cart.  & 0.345 & 0.319 & 0.698 & 0.286 & 0.0624 \\
        \textbf{Ours} & QCNet~\cite{zhou2023query} & StreamMapNet~\cite{yuan2024streammapnet} & Transformer & All  & 0.339 & 0.309 & 0.681  & \textbf{0.281}  & 0.0581 \\
        
        \midrule
        \textbf{Ours} & QCNet~\cite{zhou2023query} & StreamMapNet~\cite{yuan2024streammapnet} & 1-D U-Net & pt2pl & 0.341 & 0.318 & 0.705 & 0.375 & 0.0637 \\
        \textbf{Ours} & QCNet~\cite{zhou2023query} & StreamMapNet~\cite{yuan2024streammapnet} & 1-D U-Net & pl2pl & 0.346 & 0.308 & 0.699 & 0.328 & 0.0626 \\
        \textbf{Ours} & QCNet~\cite{zhou2023query} & StreamMapNet~\cite{yuan2024streammapnet} & 1-D U-Net & cart. & 0.345 & 0.319 & 0.698 & 0.286 & 0.0624 \\
        \textbf{Ours} & QCNet~\cite{zhou2023query} & StreamMapNet~\cite{yuan2024streammapnet} & 1-D U-Net & All & \textbf{0.336 {\color{darkpastelgreen} ($-5\%$)}} & \textbf{0.305} & \textbf{0.673 {\color{darkpastelgreen} ($-6\%$)}} & 0.309 & \textbf{0.0549 {\color{darkpastelgreen} ($-19\%$)}} \\
        \bottomrule
        
    \end{tabular}
    \vspace{-14pt}
    \label{tab:comparison_mapless_prediction}
\end{table*}
\section{Experiments}
\subsection{Dataset}
\textbf{nuScenes}~\cite{caesar2020nuscenes} is a real-world autonomous driving dataset, which includes 1,000 driving scenarios over 20 seconds. It provides agent trajectories, offline groundtruth HD maps, and sensor data. We apply the methodology in~\cite{gu2024producing} to upsample nuScenes’ data frequency to fit trajectory prediction models, which aim to predict 3-second future trajectories based on 2-second historical motion information. \textbf{NAVSIM}~\cite{dauner2024navsim} is a planning-oriented autonomous driving dataset built on OpenScene, a redistribution of nuPlan~\cite{caesar2021nuplan}. It provides eight 1920×1080 cameras and a fused LiDAR point cloud aggregated from five sensors across the current and three previous frames. The dataset is split into navtrain (1,192 training scenes) and navhard (136 evaluation scenes).
\vspace{-4pt}
\subsection{Metrics}
For trajectory prediction, We utilize standard evaluation metrics to assess the prediction performance, which include $minFDE_6$, $minADE_6$, $MR_6$. These metrics are calculated based on the best one out of 6 predicted trajectories. FDE refers to the L2 distance between the predicted trajectory and the ground-truth trajectory at the last frame, while ADE is the
average L2 error across all predicted time steps. MR (Miss Rate) refers to the proportion of predicted trajectories with an FDE larger than 2 meters. In our tables, we just use
ADE, FDE, and MR to represent $minFDE_6$, $minADE_6$ and $MR_6$ for simplicity. For end-to-end autonomous driving, we follow the official metrics of NAVSIM, Extend Predictive Driver Model Score (EPDMS) as its closed-loop planning metric as shown in equation~\ref{eq:epdms}, 
\begin{multline}
\text{EPDMS} = 
\underbrace{
\prod_{m \in \mathcal{M}_\text{pen}} \text{filter}_m(\text{agent}, \text{human})
}_{\text{penalty terms}}
\cdot \\
\underbrace{
\frac{ \sum_{m \in \mathcal{M}_\text{avg}} w_m \cdot \text{filter}_m(\text{agent}, \text{human}) }
     { \sum_{m \in \mathcal{M}_\text{avg}} w_m }
}_{\text{weighted average terms}}
\label{eq:epdms}
\end{multline}
In equation, where EPDMS integrates two sub-metrics group: $\mathcal{M}_\text{pen}=\{\text{NC}, \text{DAC}, \text{DDC}, \text{TLC}\}$ and $\mathcal{M}_\text{avg}=\{\text{TTC}, \text{EP}, \text{HC}, \text{LK}, \text{EC}\}$. No At-Fault Collision (NC), Drivable Area Compliance (DAC), Driving Direction Compl (DDC), Lane Keeping(LK),Time-to-Collision (TTC), History Comfort (HC), Extended Comfort(EC), Traffic Light Compl. (TLC) and Ego Progress (EP) to produce a comprehensive closed-loop planning score.

\subsection{Model Implementation}
For trajectory prediction, we adopt the online HD map diffusion driven QCNet~\cite{zhou2023query}. The hidden dimension is set to 128. The map polygon search radius is set to 150, with both the map polygon-to-agent and agent-to-agent search radii set to 50. The diffusion process uses a timestep of 10. The model is trained using 2 NVIDIA A100 GPUs with a batch size of 20, and 50 epochs. We employ the AdamW optimizer with an initial learning rate of $5 \times 10^{-4}$ and a weight decay of $1 \times 10^{-4}$. For planning oriented end-to-end autonomous driving, the BEV size is set 512$\times$128$\times$128 and the hidden dimension is set 512. The model is trained using 8 A100 GPUs with a batch size of 128, and 100 epochs. We employ the AdamW optimizer and a cosine learning rate decay policy. 

\subsection{Quantitative Results}
For trajectory prediction, we evaluate our approach on the nuScenes benchmark, as summarized in Table~\ref{tab:comparison_mapless_prediction}. As shown, online HD map diffusion-informed QCNet achieves SOTA performance across all metrics, improving performance by 5.1\% compared to the previous best method. Notably, it exhibits a significant reduction in Miss Rate (MR), indicating that the diffusion mechanism effectively enhances the stability and reliability of the online HD map.
For planning-oriented end-to-end autonomous driving, we further evaluate our approach on the Navsim-v2 benchmark, which includes two splits: \textit{Navtest} and \textit{Navhard}. As summarized in Table~\ref{table:result_navtest} and ~\ref{table:result_navhard}, our proposed DiffSemanticFusion achieves the best performance in the EPDMS metric on both splits, particularly on the challenging \textit{Navhard} split, where it yields an improvement of approximately 19\%. This result suggests that our method is more robust and generalizes better across diverse and complex driving scenarios.

\subsection{Ablation study}
\subsubsection{Effect of Different BEV Sizes}
We study the impact of different Bird's Eye View (BEV) feature map sizes on planning performance, with results summarized in Table~\ref{tab:bev}. The results demonstrate that the spatial resolution (height and width)  of the BEV feature map plays a more critical role than the channel dimension and has a substantial influence on planning performance.
\begin{table}[htp!]
    \scriptsize
    \centering
    \caption{\textbf{Ablation Study for BEV Sizes in Navhard}}
    \label{tab:bev}
    \begin{tabular}{l | c c c }
        \toprule
         \textbf{BEV Sizes}             &  \textbf{Stage I} $\uparrow$ & \textbf{Stage II}$\uparrow$  & \textbf{EPDMS Overall} $\uparrow$ \\
        \midrule
        256 $\times$ 32  $\times$ 32  & 57.31  & 37.43 & 22.22 \\
        512 $\times$ 32  $\times$ 32  & 58.40  & 40.33 & 23.44 \\
        512 $\times$ 64  $\times$ 64  &  71.21 & 43.26 & 30.74 \\
        512 $\times$ 128 $\times$ 128 & \textbf{72.86}  & \textbf{45.49} & 33.44 \\
        1024 $\times$ 128 $\times$ 128 & 72.56  & 45.02 & \textbf{33.56} \\
        \bottomrule 
    \end{tabular}
\end{table}
\subsubsection{Effect of Individual Components}
DiffSemanticFusion introduces a novel combination of four components: online HD map diffusion, vectorized graph embeddings, raster image embeddings, and BEV features. To evaluate their contributions, we conduct an ablation study by removing each input modality (Table~\ref{tab:ablation_component}). BEV features are the most critical, with their removal leading to a significant drop in performance. Vectorized graph and raster image embeddings further enhance spatial understanding by capturing topological and texture cues, highlighting the effectiveness of multi-modal fusion for planning tasks.

\begin{table}[htp!]
   \scriptsize
    \centering
    \caption{\textbf{Ablation Study of Individual Components in Navhard}}
    \label{tab:ablation_component}
    \begin{tabular}{c c c c c c}
        \toprule 
        \multirow{2}{*}{\makecell[c]{\textbf{Vectorized-} \\ \textbf{Graph}}}   &  
        \multirow{2}{*}{\makecell[c]{\textbf{Raster-} \\ \textbf{Image}}}      &
        \multirow{2}{*}{\makecell[c]{\textbf{BEV-} \\ \textbf{Features}}}    &
        \multicolumn{2}{c}{\textbf{EPDMS}} \\
            & & & \textbf{Stage I$\uparrow$} & \textbf{Stage II$\uparrow$} & \textbf{Overall$\uparrow$} \\
        \midrule
           $\checkmark$ & $\checkmark$ & $\checkmark$ & \textbf{72.86} & \textbf{45.49} & \textbf{33.44}\\
           $\times$ &  $\checkmark$ & $\checkmark$   & 71.69 & 43.84 &  31.43 \\
           $\checkmark$ & $\times$ & $\checkmark$    & 70.04 & 44.90 & 32.12  \\
           $\checkmark$ & $\checkmark $ & $\times$   & 56.01 & 39.33 &  20.81  \\
        \bottomrule 
    \end{tabular}
\end{table}

\subsubsection{Integration to other Models}
We integrate our proposed online HD map diffusion module into other graph-based models like VectorNet and QCNet. And online map generation method  is StreamMapNet. Table~\ref{tab:ablation_study_plugging} shows the experimental results indicating that the online HD map diffusion module can effectively improve the performance.

\begin{table}[htp!]
    \scriptsize
    \centering
    \caption{\textbf{Ablation Study for Integrating other Architectures}}
    \label{tab:ablation_study_plugging}
    \begin{tabular}{l | c c c }
        \toprule
         \textbf{Architectures}             &  \textbf{ADE} $\downarrow$ & \textbf{FDE}$\downarrow$  & \textbf{MR} $\downarrow$ \\
        \midrule
        VectorNet~\cite{Gao2020VectorNetEH} & 1.189  & 2.243 &  0.415 \\
        VectorNet + Map Diffusion           &  \textbf{1.048} & \textbf{1.921} & \textbf{0.389}  \\
        \midrule
        QCNet~\cite{Liu2023LAformerTP}      &  0.354 &  0.717  & 0.0680   \\
        QCNet + Map Diffusion               & \textbf{0.336}  & \textbf{0.673} &  \textbf{0.0549}    \\
        \bottomrule 
    \end{tabular}
\end{table}

\subsubsection{Effect of Diffusion Architecture and Coordinate System}
As shown in Table~\ref{tab:comparison_mapless_prediction}, we investigate the effects of different backbone architectures within the diffusion, including 1-D U-Net and Transformer models. In addition, we analyze the impact of injecting noise in different coordinate systems

\subsection{Qualitative results}
\subsubsection{Nuscenes Qualitative Analysis}
\begin{figure}[htp!]
  \centering
  \includegraphics[width=0.45\textwidth]{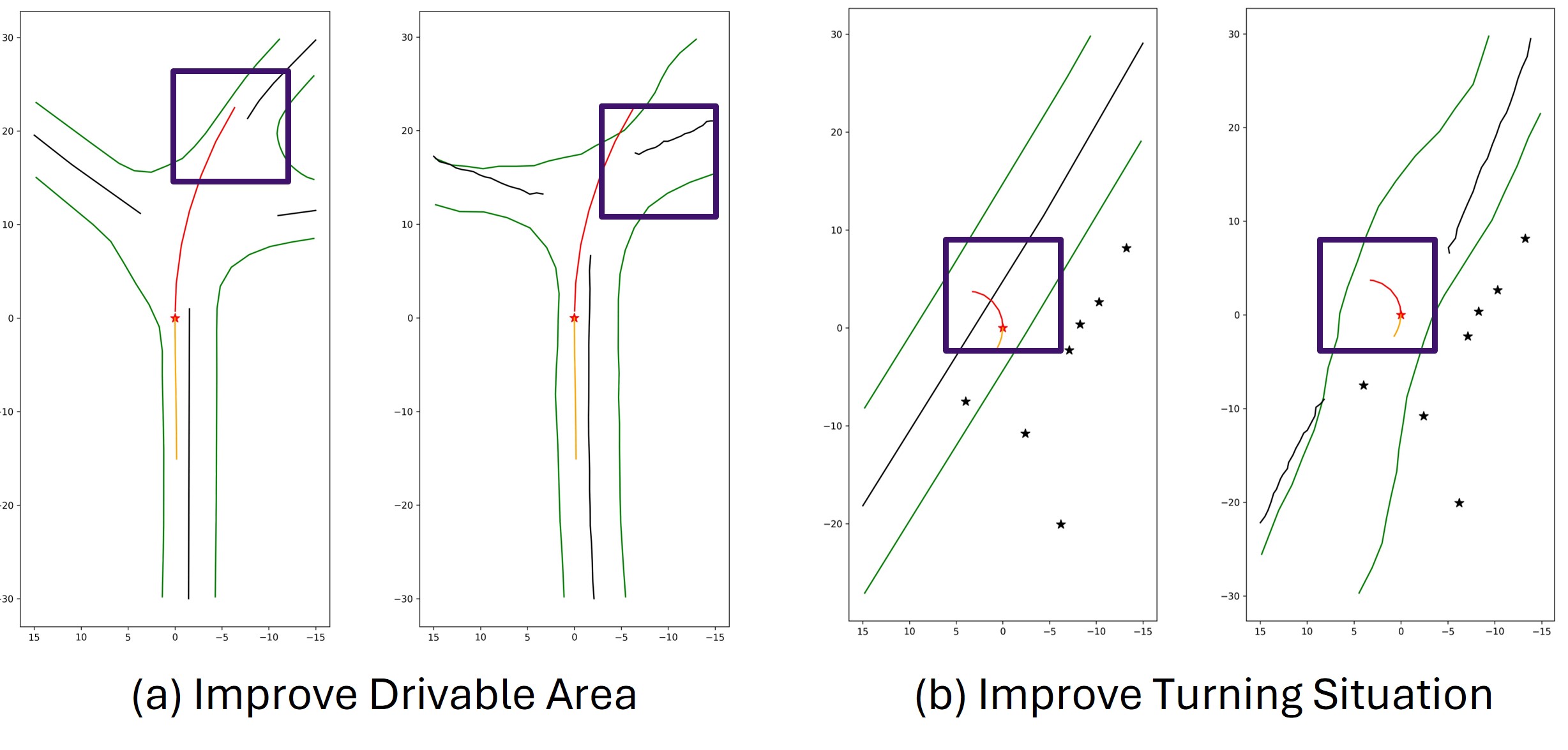} 
  \caption{nuscenes qualitative results}
    \vspace{-10pt}
  \label{fig:nusc_vis}
\end{figure}
We analyze the online map generation results of Mapless QCNet on the nuScenes dataset as shown in figure~\ref{fig:nusc_vis} and present examples that demonstrate improvements from both prediction and planning perspectives. The left side shows the ground truth, while the right side displays the online maps generated by StreamMapNet. The proposed online HD map diffusion model effectively corrects perception errors by expanding drivable areas and ensuring lane divider consistency.

\subsubsection{Navsim-v2 Qualitative Analysis}
\begin{figure}[htp!]
  \centering
  \includegraphics[width=0.45\textwidth]{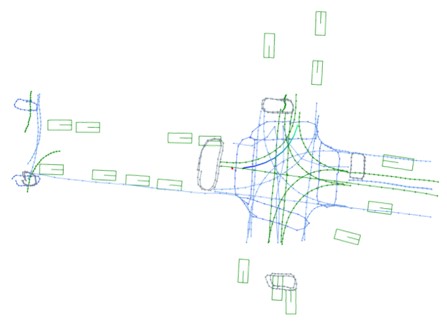} 
  \caption{navsim qualitative results}
    \vspace{-10pt}
  \label{fig:navsim_vis}
\end{figure}

To better demonstrate the benefits of the semantic raster BEV fusion in enhancing the BEV representation, we render the dynamic perception and online maps generated from intermediate sparse 4D representations in the BEV view. As shown in figure~\ref{fig:navsim_vis}, 
The trajectories planned by DiffSemanticFusion closely follow the lane centerlines, indicating accurate semantic understanding and effective planning.

\section{Conclusion}

In this work, we propose DiffSemanticFusion—a fusion framework for multimodal trajectory prediction and planning. Our approach reasons over a semantic raster–fused BEV space, enhanced by a map diffusion module that improves both the stability and expressiveness of online HD map representations. By introducing an online HD map diffusion module, our approach effectively addresses the challenges posed by noisy or incomplete online HD map inputs, significantly improving the robustness and reliability of downstream tasks. Furthermore, our semantic raster BEV fusion architecture integrates raster-based, graph-based, and BEV feature representations in a unified BEV space, leveraging their complementary strengths for more comprehensive scene understanding. We demonstrate the effectiveness of our method on both prediction and planning tasks, achieving state-of-the-art performance on the nuScenes and NAVSIM benchmarks. Notably, DiffSemanticFusion improves online HD map informed QCNet prediction by 5.1\% and boosts EPDMS in challenging NAVSIM Navhard scenarios by 15\%. These results, along with strong compatibility with other vector-based pipelines, highlight the potential of our approach to serve as a robust and generalizable solution for future autonomous driving systems. In future work, we plan to explore the integration of temporal dynamics and uncertainty modeling into the diffusion and fusion processes to further enhance long-horizon prediction and planning performance.


%%%%%%%%%%%%%%%%%%%%%%%%%%%%%%%%%%%%%%%%%%%%%%%%%%%%%%%%%%%%%%%%%%%%%%%%%%%%%%%%

%%%%%%%%%%%%%%%%%%%%%%%%%%%%%%%%%%%%%%%%%%%%%%%%%%%%%%%%%%%%%%%%%%%%%%%%%%%%%%%%



%%%%%%%%%%%%%%%%%%%%%%%%%%%%%%%%%%%%%%%%%%%%%%%%%%%%%%%%%%%%%%%%%%%%%%%%%%%%%%%%

% \clearpage

% References are important to the reader; therefore, each citation must be complete and correct. If at all possible, references should be commonly available publications.
% \newpage
% \clearpage
\bibliographystyle{IEEEtran}
\bibliography{reference}

\end{document}